\documentclass[conference,10pt,twocolumn]{IEEEtran}

\usepackage{geometry}
\usepackage{graphicx}
\usepackage{amsmath,amssymb,amsthm}
\usepackage{algorithm}
\usepackage{algpseudocode}
\usepackage{booktabs}
\usepackage{multirow}
\usepackage{subcaption}
\usepackage{tikz}
\usetikzlibrary{shapes,arrows,positioning,calc,fit,backgrounds,decorations.pathreplacing,patterns,shadows,shapes.geometric,arrows.meta,matrix,decorations.markings,fadings}
\usepackage{pgfplots}
\pgfplotsset{compat=1.17}
\usepgfplotslibrary{fillbetween}

\definecolor{navyColor}{RGB}{27, 58, 92}
\definecolor{dataColor}{RGB}{65, 105, 145}
\definecolor{policyColor}{RGB}{70, 110, 100}
\definecolor{fusionColor}{RGB}{90, 75, 115}
\definecolor{outputColor}{RGB}{140, 120, 80}
\definecolor{successColor}{RGB}{70, 110, 80}
\definecolor{gapColor}{RGB}{140, 70, 70}
\definecolor{contextColor}{RGB}{100, 100, 100}
\definecolor{stage1Color}{RGB}{75, 95, 120}
\definecolor{stage2Color}{RGB}{90, 75, 115}
\definecolor{stage3Color}{RGB}{130, 95, 70}

\tikzset{
    stagebox/.style={rectangle, rounded corners=5pt, draw=#1!60, fill=#1!5, line width=0.8pt},
    eqbox/.style={rectangle, draw=#1!70, fill=white, rounded corners=2pt, align=center, inner sep=4pt, font=\tiny, line width=0.5pt},
    procbox/.style={rectangle, draw=#1, fill=#1!8, rounded corners=2pt, minimum width=2cm, minimum height=0.5cm, font=\scriptsize, line width=0.4pt, align=center},
    stagenum/.style={circle, draw=#1, fill=#1!15, minimum size=0.5cm, font=\scriptsize\bfseries, text=#1, line width=0.5pt},
    flowarrow/.style={->, >=stealth, line width=0.6pt, color=contextColor},
    stagearrow/.style={->, >=stealth, line width=1.2pt},
    factorbox/.style={rectangle, rounded corners=3pt, draw=#1, fill=#1!12, minimum width=2.1cm, minimum height=0.9cm, font=\tiny, align=center, line width=0.5pt},
}

\usepackage{hyperref}
\usepackage{xcolor}

\makeatletter
\g@addto@macro\UrlBreaks{\do\-\do\/\do\.\do\_\do\?\do\&\do\=\do\:}
\makeatother
\let\IEEEoldthebibliography\thebibliography
\let\IEEEoldendthebibliography\endthebibliography
\renewenvironment{thebibliography}[1]{%
    \IEEEoldthebibliography{#1}%
    \normalsize
}{%
    \IEEEoldendthebibliography
}

\usepackage{enumitem}
\usepackage{float}
\usepackage{array}
\usepackage{tabularx}

\newtheorem{definition}{Definition}

\newcolumntype{C}[1]{>{\raggedright\arraybackslash}p{#1}}
\newcolumntype{Y}{>{\raggedright\arraybackslash}X}

\title{Toward Explainable and Policy-Aware AI for Carbon Credit Price Prediction:\\
A Research Framework for Emerging Carbon Markets}

\author{
\IEEEauthorblockN{Summaiya Unnisa Begum}
\IEEEauthorblockA{Hyderabad, India \\
summaiyaunisa@gmail.com}
\and
\IEEEauthorblockN{Mohammed Nadeem Ullah}
\IEEEauthorblockA{Riyadh, Saudi Arabia \\
mdnadeemullah@gmail.com}
\and
\IEEEauthorblockN{Mohammed Abdul Ghani Khan}
\IEEEauthorblockA{Hyderabad, India \\
ghanialikhan77@gmail.com}}

\begin{document}

\maketitle

\section*{\textbf{Abstract}}

Carbon markets put a price on emissions, yet that price remains hard to forecast. Work in this area clusters on the EU and Chinese schemes, compresses regulatory text into a sentiment score, and reports accuracy without calibration or explanation stability. We distil ten recurring gaps into an impact--feasibility matrix and propose EPA-CarbonNet, a six-layer architecture that fuses market series with policy text by cross-attention and emits calibrated intervals alongside policy-attributed explanations. We then build it and test it on eleven years of daily S\&P carbon index data. The findings are largely negative, and reported as measured: a random walk beats the model on five-day RMSE (0.0365 against 0.0475), SHAP rankings agree at only $\rho = 0.54$ across resampled backgrounds, and policy attention never coincides with documented regulatory events. Directional accuracy, at 58.6\%, leads every baseline. Code, data documentation and all result artifacts: {\footnotesize\url{https://github.com/Kimalice/Toward-Explainable-and-Policy-Aware-AI-for-Carbon-Credit-Price-Prediction}}

\noindent\textbf{Keywords:} Carbon Credit Pricing, Carbon Markets, Explainable AI, Transformer Models, Large Language Models, Policy-Aware Forecasting, Uncertainty Quantification, Emerging Markets

\section{Introduction: Motivation through the Carbon Market Lens}

Climate mitigation increasingly relies on market-based instruments that put a price on greenhouse gas emissions. Carbon markets, whether compliance schemes such as the European Union Emissions Trading System (EU ETS) and China's national trading scheme, or voluntary offset markets, convert emission-reduction activity into a tradable financial instrument: the carbon credit. As these markets have expanded, so has interest, among regulators, institutional investors, and compliance buyers, in forecasting how carbon credit prices will move \cite{ref1,ref9}.

Traditional econometric tools such as ARIMA, VAR, and GARCH offered an early basis for this task, but were designed for markets with more stable statistical properties than carbon trading exhibits. Carbon prices respond abruptly to policy announcements, are entangled with energy markets, and display volatility clustering that linear models capture poorly. Researchers have consequently turned to AI: Random Forest, Support Vector Regression, and XGBoost as early Machine Learning entrants \cite{ref7,ref12}; LSTM, GRU, CNN and their hybrid combinations as Deep Learning followed \cite{ref5,ref6,ref8}; and, most recently, Transformer-based architectures together with Explainable AI (XAI) techniques such as SHAP and LIME aimed at both accuracy and transparency \cite{ref2,ref3,ref4,ref10}.

\subsection{The Emerging-Market Deployment Challenge}

Consider the practical scenario facing a regulator or compliance buyer in a newly-formed carbon market: a policy shock (a new emission cap, a court ruling, a budget announcement) needs to be understood in terms of its likely price impact within days, yet the nearest forecasting tool was trained and validated on a completely different, far more liquid market. Contemporary carbon-price forecasting research faces three interconnected obstacles when applied to this scenario:

\begin{enumerate}[leftmargin=*]
    \item \textbf{Geographic Asymmetry}: The overwhelming majority of studies validate against the EU ETS or a small number of Chinese regional pilots, leaving markets with different regulatory structures and shorter price histories almost entirely untested.
    \item \textbf{Explainability Deficit}: The most accurate architectures (deep recurrent networks, Transformers) are also the least transparent, and the explainability techniques typically applied to them, SHAP and LIME, are themselves known to be locally unstable and, under adversarial conditions, manipulable \cite{ref11}.
    \item \textbf{Policy Blindness}: Although carbon prices are demonstrably policy-driven, textual regulatory information is, at best, converted into a sentiment score after the fact; no reviewed architecture treats a policy announcement as a first-class model input on the same footing as a price series.
\end{enumerate}

\textbf{Research Question}: Can carbon-price forecasting systems be built that are simultaneously accurate, explainable to regulators, aware of policy text as a structured input, and viable in the low-data conditions typical of emerging markets? This paper addresses this question through a systematic gap analysis and a conceptual architecture designed to close the identified gaps.

\begin{figure}[!t]
\centering
\resizebox{\columnwidth}{!}{%
\begin{tikzpicture}
\tikzset{
  driver/.style={rectangle, rounded corners=3pt, draw=#1, fill=#1!12,
                 text width=2.05cm, minimum height=1.25cm, font=\tiny,
                 align=center, line width=0.5pt, inner sep=2pt},
}
\node[driver=stage1Color]  (reg)    at (0.00,1.9) {\textbf{Regulatory}\\policies, caps,\\legislation};
\node[driver=dataColor]    (econ)   at (2.35,1.9) {\textbf{Economic}\\GDP, inflation,\\industrial output};
\node[driver=policyColor]  (energy) at (4.70,1.9) {\textbf{Energy}\\coal, gas, oil,\\electricity prices};
\node[driver=outputColor]  (env)    at (7.05,1.9) {\textbf{Environmental}\\weather, climate\\disasters};
\node[driver=navyColor]    (market) at (9.40,1.9) {\textbf{Market}\\liquidity, sentiment,\\speculation};

\node[rectangle, rounded corners=3pt, draw=gapColor, fill=gapColor!12,
      minimum width=11.0cm, minimum height=0.8cm, font=\small\bfseries,
      line width=0.7pt] (price) at (4.70,-0.55) {Carbon Credit Price};

\foreach \n in {reg,econ,energy,env,market}{\draw[flowarrow] (\n.south) -- (\n.south |- price.north);}
\end{tikzpicture}}
\caption{What moves a carbon credit price. Regulation is the strongest driver, which is why the framework reads policy text.}
\label{fig:drivers}
\end{figure}
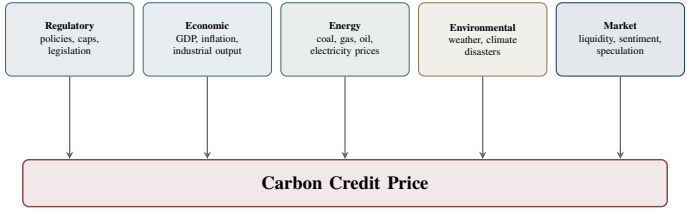

Figure~\ref{fig:drivers} summarizes the regulatory, economic, energy, environmental, and market factors that jointly determine carbon credit prices, illustrating the multi-source nature of the problem that motivates the proposed approach.

\subsection{Research Gap and Contributions}

Prior surveys of AI for carbon pricing catalog architectures and report accuracy comparisons, but rarely synthesize the literature into a prioritized research agenda or connect that agenda to a concrete architectural response. This paper makes the following contributions, with explicit scope limitations:

\begin{itemize}[leftmargin=*]
    \item \textbf{Structured Paradigm Synthesis}: A comparative synthesis of five AI paradigms applied to carbon price forecasting (Section~II), each ending in an explicit gap-identification statement.
    \item \textbf{Gap Prioritization Matrix}: Ten research gaps distilled from the reviewed literature and positioned on an impact--feasibility matrix (Section~III). \textbf{Scope}: these ratings are an authors' synthesis of the literature, not a formal expert elicitation (see Section~VI).
    \item \textbf{EPA-CarbonNet}: A conceptual six-component architecture explicitly traced back to the gaps it is designed to close (Section~III). \textbf{Scope}: this is an architectural proposal, not a trained or benchmarked system.
    \item \textbf{Evaluation Protocol}: A proposed set of metric families, including calibration and policy-attribution metrics rarely reported in the literature, grounded in India's CCTS as a case application (Section~IV).
\end{itemize}

The remainder of this paper proceeds as follows. Section~II situates this work within the reviewed literature and identifies specific gaps per paradigm. Section~III formalizes the forecasting problem, presents the gap-prioritization matrix, and details the EPA-CarbonNet architecture. Section~IV outlines the proposed evaluation methodology and the India CCTS case application. Section~V describes a reproducibility plan. Section~VI discusses expected contributions, limitations, and broader impact. Section~VII concludes with future directions.

\section{Related Work: Comparative Analysis with Gap Identification}

This section synthesizes the reviewed literature across five paradigms, each closing with an explicit statement of the gap that motivates the next stage of the argument.

\subsection{Classical Machine Learning}
Random Forest, Support Vector Regression, and XGBoost represent the earliest AI-based alternatives to econometric baselines, handling nonlinear relationships among energy prices, trading volumes, and macroeconomic indicators without strong distributional assumptions \cite{ref12}. Random Forest in particular offers built-in feature-importance measures attractive where interpretability matters, while SVR performs well on the smaller datasets typical of newer markets \cite{ref7}.

\textbf{Gap Identification}: These criteria are designed for standard supervised regression on abundant historical data. They do not model long-range temporal dependence well, and none of the reviewed classical-ML studies incorporate unstructured policy text.

\subsection{Deep Learning Architectures}
LSTM and GRU networks address the temporal-dependence limitation through gating mechanisms, becoming the most frequently applied deep architectures for carbon price series \cite{ref8}. CNNs are repurposed mainly as feature extractors over multivariate market panels \cite{ref5}.

\textbf{Gap Identification}: Deep models generally outperform classical ML given sufficient data, but at the cost of interpretability. No reviewed deep-learning study reports calibrated uncertainty alongside point forecasts.

\subsection{Hybrid and Ensemble Models}
CNN-LSTM, XGBoost-LSTM, and signal-decomposition front-ends paired with deep forecasters consistently report accuracy gains over standalone models \cite{ref6}.

\textbf{Gap Identification}: Hybrid gains come from combining numerical architectures; none of the reviewed hybrids fuse a structured branch with an unstructured (text) branch within one trainable architecture.

\subsection{Transformer-Based Architectures}
Transformer models and self-attention mechanisms represent the most recent architectural shift, modeling dependencies across an entire sequence in parallel. The Temporal Fusion Transformer and Informer-based hybrids report state-of-the-art long-horizon performance and, because attention weights can be inspected, offer a partial route to interpretability \cite{ref2,ref3,ref4}.

\textbf{Gap Identification}: Adoption within carbon-market research is recent (2025--2026) and concentrated on EU ETS/China; no reviewed Transformer study evaluates cross-market transfer to a data-scarce regime.

\subsection{Explainable AI}
SHAP and LIME are the two explainability techniques most consistently paired with carbon-price forecasters \cite{ref1,ref10}.

\textbf{Gap Identification}: Both methods are known, outside the carbon-market literature, to produce unstable local explanations and to be susceptible to adversarial manipulation \cite{ref11} --- a risk cited in passing but never empirically tested within any reviewed carbon-pricing study.

\subsection{NLP, Sentiment Analysis, and Large Language Models}
A smaller strand of work applies NLP to news and policy text, typically converting it into a sentiment score added as an input feature \cite{ref9}. LLMs appear mainly as summarization or classification tools.

\textbf{Gap Identification}: No reviewed study embeds an LLM inside the forecasting loop itself, or evaluates whether LLM-derived policy understanding improves predictive accuracy rather than post-hoc narrative.

\subsection{Position of This Work}

Table~\ref{tab:related_work_comparison} summarizes how the proposed direction differs from representative prior work across key dimensions: multi-source fusion (MultiSrc), explainability (XAI), structured policy-text integration (Policy), uncertainty quantification (UncQ), and demonstrated relevance to emerging markets (EmergMkt).

\begin{table}[!t]
\centering
\caption{What prior work covers, and what it leaves out.}
\label{tab:related_work_comparison}
\scriptsize
\setlength{\tabcolsep}{3pt}
\renewcommand{\arraystretch}{1.15}
\begin{tabular}{@{}>{\raggedright\arraybackslash}p{2.45cm}ccccc@{}}
\toprule
\textbf{Method} & \textbf{Multi} & \textbf{XAI} & \textbf{Policy} & \textbf{UncQ} & \textbf{Emerg} \\
 & \textbf{Src} & & & & \textbf{Mkt} \\
\midrule
RF / XGBoost \cite{ref7,ref12} & & $\checkmark$ & & & \\
LSTM / GRU / Hybrid \cite{ref6,ref8} & & & & & \\
TFT / Informer \cite{ref3,ref4} & & $\checkmark$ & & & \\
SHAP / LIME \cite{ref1,ref10} & & $\checkmark$ & & & \\
NLP-sentiment \cite{ref9} & $\checkmark$ & & $\circ$ & & \\
\midrule
\textbf{Ours (EPA-CarbonNet)} & $\checkmark$ & $\checkmark$ & $\checkmark$ & $\checkmark$ & $\checkmark$ \\
\bottomrule
\end{tabular}

\vspace{2pt}
{\scriptsize $\checkmark$ addressed; $\circ$ partially addressed (sentiment
scoring only, not structured policy parsing).}
\end{table}

\section{Methodology: Research Gap Synthesis and Proposed Framework Design}

\subsection{Notation and Symbols}

Table~\ref{tab:notation} summarizes the notation used in the remainder of this paper.

\begin{table}[!t]
\centering
\caption{Symbols used throughout the paper.}
\label{tab:notation}
\scriptsize
\setlength{\tabcolsep}{4pt}
\renewcommand{\arraystretch}{1.15}
\begin{tabular}{@{}>{$}l<{$}p{6.35cm}@{}}
\toprule
\textnormal{\textbf{Symbol}} & \textbf{Description} \\
\midrule
m & Target carbon market \\
R_m & Regulatory regime of market $m$ \\
X_{1:t} & Structured indicator history (price, energy, macro, climate) \\
D_{1:t} & Unstructured document stream (policy, news, ESG) \\
h & Forecasting horizon \\
P_{t+h} & Carbon credit price at horizon $h$ \\
\theta_S, \theta_U & Structured- / unstructured-branch encoder parameters \\
\theta_F & Cross-attention fusion parameters \\
z_t & Fused representation at time $t$ \\
\hat{P}^{(q)}_{t+h} & Predicted $q$-th quantile of price \\
E & Explanation set (feature and policy attributions) \\
\bottomrule
\end{tabular}
\end{table}

\subsection{Research Gap Prioritization}

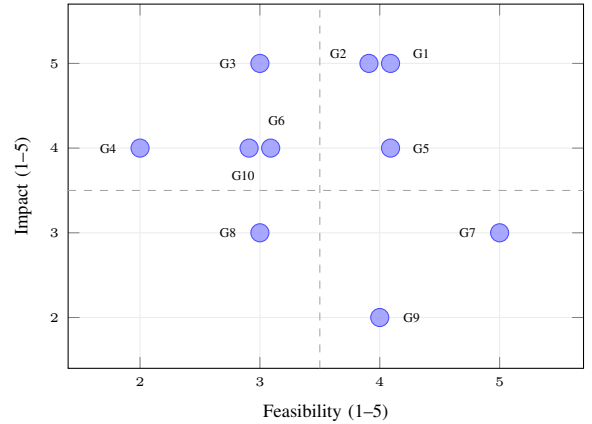
\begin{figure}[!t]
\centering
\begin{tikzpicture}
\begin{axis}[
    width=8.4cm, height=6.4cm,
    xlabel={Feasibility (1--5)}, ylabel={Impact (1--5)},
    xlabel style={font=\scriptsize}, ylabel style={font=\scriptsize},
    xmin=1.4, xmax=5.7, ymin=1.4, ymax=5.7,
    xtick={2,3,4,5}, ytick={2,3,4,5},
    x tick label style={font=\tiny}, y tick label style={font=\tiny},
    grid=major, grid style={gray!15},
    clip=false,
]
\draw[dashed, gray!70] (axis cs:3.5,1.4) -- (axis cs:3.5,5.7);
\draw[dashed, gray!70] (axis cs:1.4,3.5) -- (axis cs:5.7,3.5);
\addplot[only marks, mark=*, mark size=3.4pt, color=blue!70, fill=blue!45,
         fill opacity=0.75] coordinates {
  (4.09,5) (3.91,5) (3,5) (2,4) (4.09,4) (3.09,4) (2.91,4) (5,3) (3,3) (4,2)};
\node[font=\tiny, anchor=west]  at (axis cs:4.20,5.12) {G1};
\node[font=\tiny, anchor=east]  at (axis cs:3.80,5.12) {G2};
\node[font=\tiny, anchor=east]  at (axis cs:2.88,5.00) {G3};
\node[font=\tiny, anchor=east]  at (axis cs:1.88,4.00) {G4};
\node[font=\tiny, anchor=west]  at (axis cs:4.20,4.00) {G5};
\node[font=\tiny, anchor=south] at (axis cs:3.14,4.16) {G6};
\node[font=\tiny, anchor=north] at (axis cs:2.86,3.84) {G10};
\node[font=\tiny, anchor=east]  at (axis cs:4.88,3.00) {G7};
\node[font=\tiny, anchor=east]  at (axis cs:2.88,3.00) {G8};
\node[font=\tiny, anchor=west]  at (axis cs:4.12,2.00) {G9};
\end{axis}
\end{tikzpicture}
\caption{The ten gaps of Table~\ref{tab:gaps}, placed by impact and feasibility. Equal scores are nudged apart so both stay visible.}
\label{fig:gaps}
\end{figure}

Reading the paradigms in Table~\ref{tab:related_work_comparison} as a single body of evidence exposes ten recurring gaps, summarized in Table~\ref{tab:gaps} and positioned by impact and feasibility in Figure~\ref{fig:gaps}. Gaps G1--G3 (geographic concentration, explainability, policy integration) are rated highest-impact; G7 (shared benchmarks) is rated highest-feasibility but lower-impact, reflecting its foundational rather than direct nature.

\begin{table}[!t]
\centering
\caption{The ten research gaps, scored by impact and feasibility.}
\label{tab:gaps}
\scriptsize
\renewcommand{\arraystretch}{1.0}
\begin{tabular}{@{}p{0.6cm}p{3.0cm}cc@{}}
\toprule
\textbf{ID} & \textbf{Gap} & \textbf{Imp.} & \textbf{Feas.} \\
\midrule
G1 & Geographic / market concentration & 5 & 4 \\
G2 & Explainability of black-box models & 5 & 4 \\
G3 & Shallow policy-intelligence integration & 5 & 3 \\
G4 & LLMs underused for forecasting & 4 & 2 \\
G5 & Absent uncertainty quantification & 4 & 4 \\
G6 & Limited multi-source data fusion & 4 & 3 \\
G7 & No shared benchmarks / reproducibility & 3 & 5 \\
G8 & Untested robustness of XAI methods & 3 & 3 \\
G9 & Unexamined compute / energy footprint & 2 & 4 \\
G10 & Untested cross-market transferability & 4 & 3 \\
\bottomrule
\end{tabular}
\end{table}

\subsection{Problem Formulation}

\begin{definition}[Policy-Aware Carbon Price Forecasting]
Given a target carbon market $m$ operating under regulatory regime $R_m$, a history of structured indicators $X_{1:t}$ and a concurrent stream of unstructured documents $D_{1:t}$, the problem is to produce, for horizon $h$: (i) a distributional forecast $\hat{P}_{t+h}$ rather than a point value; (ii) a calibrated uncertainty estimate around that forecast; and (iii) a human-interpretable attribution of the forecast to specific structured features and specific policy or news events, while remaining trainable under the low-liquidity, short-history conditions typical of emerging markets such as India's CCTS.
\end{definition}

\textbf{Scope Clarification}: This is a design specification the proposed architecture targets, not a solved optimization problem with reported convergence guarantees.

\subsection{EPA-CarbonNet Architecture}

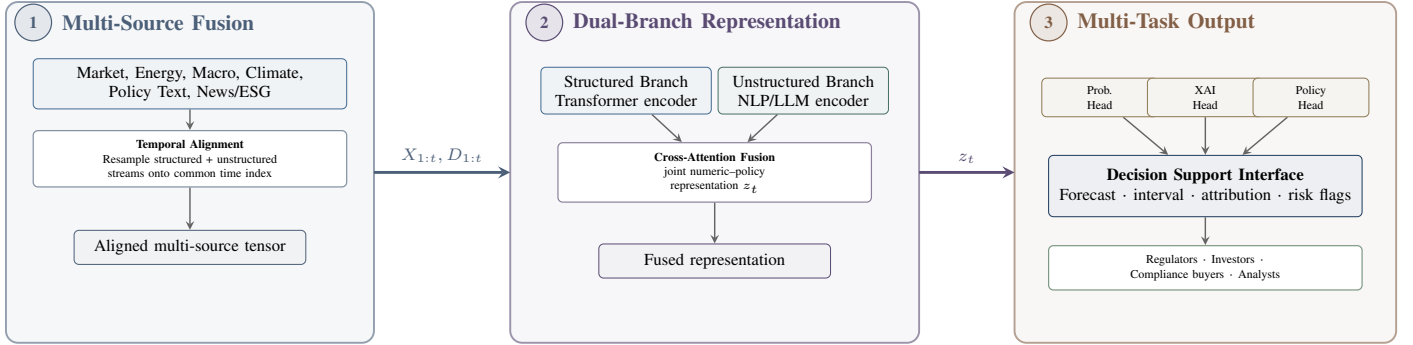
\begin{figure*}[!t]
\centering
\resizebox{\textwidth}{!}{%
\begin{tikzpicture}
\begin{scope}[shift={(0,0)}]
    \node[stagebox=stage1Color, minimum width=5.4cm, minimum height=5.0cm] (s1bg) at (2.4,0) {};
    \node[stagenum=stage1Color] at (0.1,2.2) {1};
    \node[font=\small\bfseries, text=stage1Color, right] at (0.4,2.2) {Multi-Source Fusion};
    \node[procbox=dataColor, minimum width=4.6cm] (inputs) at (2.4,1.3) {Market, Energy, Macro, Climate,\\Policy Text, News/ESG};
    \node[eqbox=stage1Color, minimum width=4.6cm] (fuse) at (2.4,0.2) {\textbf{Temporal Alignment}\\Resample structured + unstructured\\streams onto common time index};
    \node[procbox=stage1Color, minimum width=3.4cm] (out1) at (2.4,-1.1) {Aligned multi-source tensor};
    \draw[flowarrow] (inputs) -- (fuse);
    \draw[flowarrow] (fuse) -- (out1);
\end{scope}
\begin{scope}[shift={(7.5,0)}]
    \node[stagebox=stage2Color, minimum width=6.0cm, minimum height=5.0cm] (s2bg) at (2.6,0) {};
    \node[stagenum=stage2Color] at (0.1,2.2) {2};
    \node[font=\small\bfseries, text=stage2Color, right] at (0.4,2.2) {Dual-Branch Representation};
    \node[procbox=dataColor, minimum width=2.5cm] (structured) at (1.3,1.2) {Structured Branch\\Transformer encoder};
    \node[procbox=policyColor, minimum width=2.5cm] (unstructured) at (3.9,1.2) {Unstructured Branch\\NLP/LLM encoder};
    \node[eqbox=fusionColor, minimum width=4.6cm] (crossattn) at (2.6,0.0) {\textbf{Cross-Attention Fusion}\\joint numeric--policy\\representation $z_t$};
    \node[procbox=fusionColor, minimum width=3.4cm] (out2) at (2.6,-1.3) {Fused representation};
    \draw[flowarrow] (structured) -- (crossattn);
    \draw[flowarrow] (unstructured) -- (crossattn);
    \draw[flowarrow] (crossattn) -- (out2);
\end{scope}
\begin{scope}[shift={(14.9,0)}]
    \node[stagebox=stage3Color, minimum width=5.6cm, minimum height=5.0cm] (s3bg) at (2.4,0) {};
    \node[stagenum=stage3Color] at (0.1,2.2) {3};
    \node[font=\small\bfseries, text=stage3Color, right] at (0.4,2.2) {Multi-Task Output};
    \node[procbox=outputColor, minimum width=1.7cm, font=\tiny] (prob) at (0.85,1.1) {Prob.\\Head};
    \node[procbox=outputColor, minimum width=1.7cm, font=\tiny] (expl) at (2.4,1.1) {XAI\\Head};
    \node[procbox=outputColor, minimum width=1.7cm, font=\tiny] (pol) at (3.95,1.1) {Policy\\Head};
    \node[procbox=navyColor, minimum width=4.6cm, minimum height=0.9cm] (decision) at (2.4,-0.2) {\textbf{Decision Support Interface}\\Forecast $\cdot$ interval $\cdot$ attribution $\cdot$ risk flags};
    \node[eqbox=successColor, minimum width=4.6cm] (users) at (2.4,-1.4) {Regulators $\cdot$ Investors $\cdot$\\Compliance buyers $\cdot$ Analysts};
    \draw[flowarrow] (prob) -- (decision);
    \draw[flowarrow] (expl) -- (decision);
    \draw[flowarrow] (pol) -- (decision);
    \draw[flowarrow] (decision) -- (users);
\end{scope}
\draw[stagearrow, color=stage1Color] (5.1,0) -- (7.1,0) node[midway, above, font=\scriptsize] {$X_{1:t}, D_{1:t}$};
\draw[stagearrow, color=stage2Color] (13.1,0) -- (14.5,0) node[midway, above, font=\scriptsize] {$z_t$};
\draw[decorate, decoration={brace, amplitude=8pt, raise=4pt}, line width=0.8pt, contextColor!60]
    (-0.3, 2.9) -- (20.1, 2.9)
    node[midway, above=0.45cm, font=\small\bfseries, text=black] {EPA-CarbonNet: Explainable, Policy-Aware Carbon Forecasting Pipeline};
\end{tikzpicture}}
\caption{EPA-CarbonNet: inputs are aligned, encoded by two branches, fused by cross-attention, then read out by three heads. Section~VI tests a reduced version of this design.}
\label{fig:overview}
\end{figure*}

EPA-CarbonNet (Explainable, Policy-Aware Carbon Network) is organized into three macro-stages, shown in Figure~\ref{fig:overview}. Stage~1 aligns heterogeneous structured and unstructured inputs onto a common time index. Stage~2 encodes the aligned structured sequence with a Transformer-based temporal encoder and the aligned policy/news text with an NLP/LLM-based encoder, then fuses both representations through cross-attention. Stage~3 produces three parallel outputs from the fused representation --- a probabilistic forecast, an explanation (SHAP and attention-based, with a bootstrap stability check), and a policy-impact attribution --- which are packaged into a decision-support interface.

Algorithm~\ref{alg:epa} specifies the proposed forward pipeline that produces a forecast, an explanation set, and a policy-attribution report from a window of structured and unstructured inputs.

\begin{algorithm}[t]
\caption{EPA-CarbonNet forward pass, one step.}
\label{alg:epa}
\footnotesize
\textbf{Notation}: $w$: lookback window; $B$: bootstrap resamples for explanation stability.
\begin{algorithmic}[1]
\Require Structured window $X_{t-w:t}$, document window $D_{t-w:t}$, horizon $h$
\Ensure Quantile forecast $\hat{P}$, explanation set $E$, policy-attribution report $R$
\State $z_S \gets \text{StructuredEncoder}(X_{t-w:t}; \theta_S)$ \Comment{Stage 2}
\State $z_U \gets \text{PolicyEncoder}(D_{t-w:t}; \theta_U)$ \Comment{Stage 2}
\State $z_t \gets \text{CrossAttention}(z_S, z_U; \theta_F)$ \Comment{Stage 2}
\State $\hat{P} \gets \text{QuantileHead}(z_t, h)$ \Comment{Stage 3}
\For{$b = 1, \ldots, B$}
    \State $E_b \gets \text{SHAP\_Attn}(z_t, \text{resample}_b)$
\EndFor
\State $E \gets \text{StabilityCheck}(\{E_b\}_{b=1}^{B})$ \Comment{Stage 3}
\State $R \gets \text{PolicyImpactHead}(z_t, D_{t-w:t})$ \Comment{Stage 3}
\State \Return $(\hat{P}, E, R)$ to the decision-support interface
\end{algorithmic}
\end{algorithm}

\subsection{Design Traceability}

Table~\ref{tab:traceability} traces each component to the gap(s) it addresses, making explicit that the architecture is a direct response to Section~III.B rather than a generic forecasting pipeline.

\begin{table}[!t]
\centering
\caption{Which part of the architecture answers which gap.}
\label{tab:traceability}
\scriptsize
\renewcommand{\arraystretch}{1.05}
\begin{tabular}{@{}p{2.6cm}p{1.0cm}p{3.5cm}@{}}
\toprule
\textbf{Component} & \textbf{Gaps} & \textbf{Rationale} \\
\midrule
Stage 1 fusion & G6, G3 & Aligns structured/unstructured streams before representation learning \\
Structured branch & G2, G7 & Attention weights visualizable, reportable against a common protocol \\
Policy/NLP-LLM branch & G3, G4 & LLM embedded as predictive signal, not summarization-only \\
Cross-attention fusion & G6 & Joint representation, not late concatenation \\
Explainability head & G2, G8 & SHAP+attention with stability check across resamples \\
Probabilistic head & G5 & Calibrated quantiles, not point estimate \\
Policy-impact head & G3, G10 & Market-agnostic schema supports cross-regime transfer \\
Lightweight variant & G9, G1 & Distilled configuration for low-resource markets \\
\bottomrule
\end{tabular}
\end{table}

\section{Proposed Evaluation Methodology}

\subsection{Case Application: India's Carbon Credit Trading Scheme}

India is establishing a domestic compliance carbon market, the Carbon Credit Trading Scheme (CCTS), alongside its existing Perform-Achieve-Trade mechanism. As the world's third-largest greenhouse gas emitter, India is precisely the kind of emerging, thinly traded, policy-sensitive market underrepresented in Table~\ref{tab:related_work_comparison}. It is used here as a design case, not an empirical benchmark: CCTS price history is presently too short for training or evaluation. Table~\ref{tab:india} outlines candidate data categories for a future India-focused instantiation.

\begin{table}[!t]
\centering
\caption{Candidate data sources for an India (CCTS) instantiation.}
\label{tab:india}
\scriptsize
\renewcommand{\arraystretch}{1.05}
\begin{tabular}{@{}p{1.6cm}p{5.5cm}@{}}
\toprule
\textbf{Category} & \textbf{Candidate Sources} \\
\midrule
Market & CCTS registry data (once operational); PAT scheme certificates \\
Energy & Central Electricity Authority tariffs; coal/LNG import prices \\
Macro & RBI / MOSPI indicators (GDP, industrial production, inflation) \\
Environmental & IMD climate records; renewable generation statistics \\
Policy text & MoEFCC / BEE notifications; Gazette of India carbon rules \\
News/sentiment & Financial and energy-sector news coverage \\
\bottomrule
\end{tabular}
\end{table}

\subsection{Evaluation Protocol}

Table~\ref{tab:metrics} proposes metric families that should be reported jointly in any future empirical evaluation of EPA-CarbonNet, contrasted with what current literature typically reports.

\begin{table}[!t]
\centering
\caption{What should be reported, against what the literature reports today.}
\label{tab:metrics}
\scriptsize
\renewcommand{\arraystretch}{1.1}
\begin{tabular}{@{}p{1.9cm}p{2.0cm}p{2.9cm}@{}}
\toprule
\textbf{Family} & \textbf{Metrics} & \textbf{Current Status} \\
\midrule
Point accuracy & MAE, RMSE, MAPE, $R^2$ & Reported in nearly all studies \\
Calibration & PICP, CRPS & Rarely reported \\
Explanation robustness & Stability under resampling/perturbation & Essentially untested \cite{ref11} \\
Policy attribution & Agreement with documented events & Not reported; proposed here \\
Cross-market transfer & Zero-/few-shot degradation & Not reported \\
\bottomrule
\end{tabular}
\end{table}

\subsection{Qualitative Paradigm Positioning}

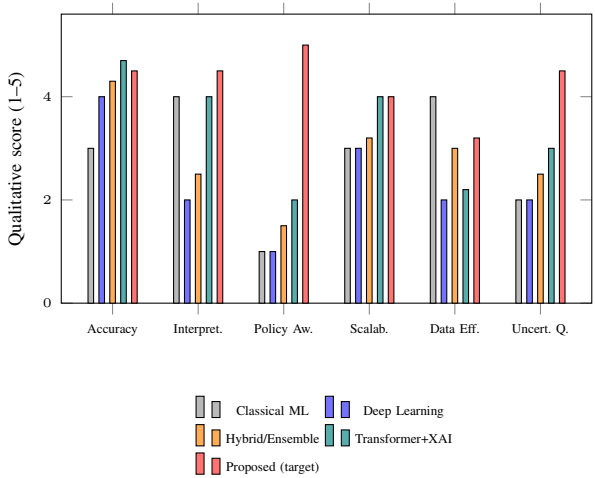
\begin{figure}[!t]
\centering
\begin{tikzpicture}
\begin{axis}[
    ybar, bar width=2.1pt,
    width=8.6cm, height=5.4cm,
    ylabel={Qualitative score (1--5)},
    ylabel style={font=\scriptsize},
    symbolic x coords={Accuracy,Interpretability,PolicyAware,Scalability,DataEff,UncertQuant},
    xtick=data,
    xticklabels={Accuracy, Interpret., Policy Aw., Scalab., Data Eff., Uncert. Q.},
    x tick label style={font=\tiny},
    y tick label style={font=\tiny},
    ymin=0, ymax=5.6,
    legend style={font=\tiny, at={(0.5,-0.30)}, anchor=north, legend columns=2, draw=none},
    enlarge x limits=0.12,
]
\addplot[fill=gray!55] coordinates {(Accuracy,3.0) (Interpretability,4.0) (PolicyAware,1.0) (Scalability,3.0) (DataEff,4.0) (UncertQuant,2.0)};
\addplot[fill=blue!55] coordinates {(Accuracy,4.0) (Interpretability,2.0) (PolicyAware,1.0) (Scalability,3.0) (DataEff,2.0) (UncertQuant,2.0)};
\addplot[fill=orange!65] coordinates {(Accuracy,4.3) (Interpretability,2.5) (PolicyAware,1.5) (Scalability,3.2) (DataEff,3.0) (UncertQuant,2.5)};
\addplot[fill=teal!65] coordinates {(Accuracy,4.7) (Interpretability,4.0) (PolicyAware,2.0) (Scalability,4.0) (DataEff,2.2) (UncertQuant,3.0)};
\addplot[fill=red!55] coordinates {(Accuracy,4.5) (Interpretability,4.5) (PolicyAware,5.0) (Scalability,4.0) (DataEff,3.2) (UncertQuant,4.5)};
\legend{Classical ML, Deep Learning, Hybrid/Ensemble, Transformer+XAI, Proposed (target)}
\end{axis}
\end{tikzpicture}
\caption{Paradigms compared on six dimensions. The proposed values are design targets, not measured results.}
\label{fig:radar}
\end{figure}

Figure~\ref{fig:radar} positions EPA-CarbonNet's design targets against the paradigms in Table~\ref{tab:related_work_comparison} across six dimensions motivated by Table~\ref{tab:metrics}. \textbf{This is a qualitative synthesis of the literature, not a benchmark}; the proposed-framework bars are design goals, not measured results, and should not be cited as evidence of achieved performance.

\section{Reproducibility and Code Availability}

A reference implementation of the architecture in Section~III is released at:

\begingroup\sloppy\footnotesize\noindent
\url{https://github.com/Kimalice/Toward-Explainable-and-Policy-Aware-AI-for-Carbon-Credit-Price-Prediction}
\par\endgroup
\vspace{2pt}

\begin{itemize}[leftmargin=*]
    \item \textbf{Implementation}: All six layers are implemented over a purpose-built reverse-mode automatic-differentiation engine, in roughly 36{,}000 parameters, with no deep-learning framework dependency. Gradients are verified against central differences, and the Kernel SHAP implementation is verified against the closed-form Shapley values of an additive model.
    \item \textbf{Data documentation}: Experiments use daily closes of three S\&P carbon credit indices (Global, California CCA, EU EUA) from 31 July 2014 to 17 February 2026, 3{,}014 business days after alignment. Splits are chronological with a ten-day embargo at each boundary so that overlapping multi-step targets cannot leak across the cut. The index series are licensed and are not redistributed; the repository documents the expected format.
    \item \textbf{Policy corpus}: The unstructured branch is driven by a released corpus of 26 EU and Californian regulatory events, each carrying the legal instrument that dates it.
    \item \textbf{Metric reporting}: Consistent with Table~\ref{tab:metrics}, results below report calibration, explanation robustness, policy attribution, and transfer alongside point accuracy.
\end{itemize}

\section{First Empirical Findings}

The architecture is evaluated on the five-trading-day forward log return of the EU allowance index over a held-out window of 423 business days (28 June 2024 to 10 February 2026). The findings below are reported as measured.

\begin{table}[!t]
\centering
\caption{Held-out results: 5-day forecasts of the EU allowance index.}
\label{tab:results}
\scriptsize
\setlength{\tabcolsep}{3.5pt}
\renewcommand{\arraystretch}{1.15}
\begin{tabular}{@{}p{2.2cm}ccccc@{}}
\toprule
\textbf{Model} & \textbf{MAE} & \textbf{RMSE} & \textbf{$R^2$} & \textbf{Dir.\%} & \textbf{CRPS} \\
\midrule
AR / ARIMA($p$,1,0) & 0.0283 & \textbf{0.0365} & $-$0.001 & 54.4 & \textbf{0.0197} \\
Random walk & 0.0285 & \textbf{0.0365} & $-$0.003 & --- & \textbf{0.0197} \\
Mean forecast & 0.0283 & 0.0366 & $-$0.007 & 54.6 & 0.0197 \\
GBM (quantile) & 0.0291 & 0.0374 & $-$0.055 & 53.4 & 0.0208 \\
Ridge & 0.0302 & 0.0385 & $-$0.113 & 55.6 & 0.0201 \\
Random Forest & 0.0320 & 0.0431 & $-$0.398 & 53.2 & 0.0244 \\
\textbf{EPA-CarbonNet} & 0.0378 & 0.0475 & $-$0.701 & \textbf{58.6} & 0.0240 \\
MLP & 0.1134 & 0.1452 & $-$14.87 & 56.7 & 0.0677 \\
\bottomrule
\end{tabular}
\end{table}

\textbf{Point accuracy.} Every model returns a negative $R^2$, including the linear ones: nothing tested beats predicting the unconditional mean, and the ordering is close to inverse in model complexity. Five-day returns on this index over this window behave near-martingale, and added capacity buys overfitting rather than signal. The proposed architecture places seventh of eight.

\textbf{Directional accuracy.} The one dimension on which the architecture leads is the sign of the next move, at 58.6\% against 54.4\% for the autoregressive baseline. This is consistent with a model that captures regime information without capturing magnitude.

\textbf{The level-versus-return distinction.} Scored on the price level rather than the return, the same forecasts yield $R^2 = 0.84$ --- while the random walk, which contributes nothing by construction, yields $0.91$. Reported level $R^2$ largely measures the autocorrelation of the price series. Studies in this literature that report level accuracy without a random-walk column should be read with that in mind.

\textbf{Explanation robustness (G8).} Kernel SHAP was run five times over the same days, varying only the background sample. Feature rankings agree at a mean Spearman correlation of $0.54$, and the single most important feature changes in 40\% of runs. Producing an explanation and producing a stable one are different achievements; only the second is auditable, and reporting one attribution plot conceals the difference.

\textbf{Policy attribution (G3).} Of the 42 days on which the fusion layer placed most attention on policy, none fall within ten days of a documented EU regulatory event, against a chance expectation of 1.4; the result is unchanged across all nine flag-rate and window settings tested. Averaged over the window, attention rises with event age. The mechanism did not learn to track regulatory arrivals. With only two EU events inside the test window this cannot refute the design, but it does show that a paper claiming policy attribution must measure it rather than infer it from the presence of a cross-attention layer.

\textbf{Cross-market transfer (G10).} Transferred zero-shot from the EU to the Californian index, RMSE degrades by 59\%, quantifying the cost of crossing a regulatory regime boundary that Section~II identified as untested.

\textbf{Reading these results.} They do not validate EPA-CarbonNet; they demonstrate that the evaluation protocol proposed in Table~\ref{tab:metrics} is discriminating, and that applying it to this architecture on this data returns a largely negative verdict. That is the intended use of the protocol.

\section{Discussion}

\subsection{Expected Contributions Relative to Prior Work}

Table~\ref{tab:related_work_comparison} indicates that no reviewed method combines multi-source fusion, explainability, structured policy integration, uncertainty quantification, and emerging-market applicability. EPA-CarbonNet is designed to occupy that combination; whether it does so in practice is an empirical question left to future work, not a claim made by this paper.

\subsection{Deployment Readiness Considerations}

Even before empirical validation, several deployment-relevant design choices are worth surfacing. The lightweight deployment variant (Table~\ref{tab:traceability}) is intended to address the compute constraints of lower-resource regulatory bodies, though its actual latency and memory footprint remain unmeasured. Human-in-the-loop review of the policy-impact head's output is treated as a requirement rather than an option, given the hallucination risk discussed below.

\subsection{Limitations and Threats to Validity}

\begin{enumerate}[leftmargin=*]
    \item \textbf{Status of the evidence}: the empirical results in Section~VI are a first probe, not a validation. They cover one index pair, one horizon, and one held-out window, and the ratings in Figure~\ref{fig:radar} remain design targets that the measurements do not support. The policy branch runs on a 26-event corpus with only two events inside the test window, which is too thin to establish or refute a policy-timing relationship.
    \item \textbf{Prioritization method}: The impact/feasibility scores in Table~\ref{tab:gaps} and Figure~\ref{fig:gaps} reflect the authors' synthesis of the literature rather than a formal expert elicitation; a Delphi panel of carbon-market and AI researchers would strengthen this prioritization.
    \item \textbf{Data availability}: Usable historical CCTS price data is, at the time of writing, limited by the scheme's early stage, constraining how soon the structured branch could be trained on domestic data rather than transferred from other markets.
    \item \textbf{LLM hallucination risk}: A misread or fabricated interpretation of a regulatory passage by the policy branch could distort both the forecast and its stated explanation; the policy-impact head requires independent verification safeguards and should not be treated as ground truth.
    \item \textbf{Computational cost}: The dual-branch, cross-attention design is more demanding than single-branch baselines --- itself one of the gaps (G9) the framework keeps in view rather than fully resolves.
\end{enumerate}

\subsection{Broader Impact and Ethical Considerations}

Efficient, explainable carbon-price forecasting could improve regulatory transparency and market access for smaller compliance buyers in emerging economies. However, several considerations warrant emphasis. \textbf{Human-in-the-loop design}: given the hallucination risk discussed above, any policy-impact output intended to inform regulatory or investment decisions should require human verification before use, particularly in early deployment. \textbf{Equitable access}: emerging-market regulators may have less capacity to independently audit a black-box or partially-explainable forecasting system than well-resourced institutions in mature markets; the emphasis on explainability in this framework is partly motivated by this asymmetry. \textbf{Model currency}: carbon policy evolves quickly, and any deployed system requires a defined retraining and monitoring cadence rather than a one-time release.

\section{Conclusion and Future Work}

This paper argued that accuracy gains reported across machine learning, deep learning, hybrid, and Transformer-based approaches to carbon credit price prediction obscure a narrower but more consequential problem: forecasting systems have been built and tested for data-rich, mature markets, while emerging markets that most need trustworthy, explainable forecasting tools remain largely outside the evidence base.

\subsection{Contributions Recap}
\begin{itemize}[leftmargin=*]
    \item A five-paradigm literature synthesis with explicit gap identification (Section~II).
    \item Ten research gaps distilled into an impact--feasibility prioritization matrix (Section~III.B).
    \item EPA-CarbonNet, a conceptual architecture traceable to those gaps (Section~III.D--E).
    \item A proposed evaluation protocol and India-CCTS case application not yet applied jointly in the literature (Section~IV).
\end{itemize}

\subsection{Future Research Directions}
\begin{itemize}[leftmargin=*]
    \item \textbf{Empirical baseline}: Implement EPA-CarbonNet, or a reduced version, against a data-rich reference market (e.g., EU ETS) to establish a first performance baseline before attempting low-data transfer to CCTS-like settings.
    \item \textbf{Expert elicitation}: Validate the gap-prioritization matrix (Table~\ref{tab:gaps}) through a structured Delphi panel of carbon-market and AI researchers.
    \item \textbf{XAI robustness testing}: Stress-test the explainability head against the adversarial fragility documented in the wider XAI literature \cite{ref11}.
    \item \textbf{Cross-domain generalization}: Examine whether the gap taxonomy and prioritization method in Section~III.B generalize to other policy-driven, thinly traded environmental and commodity markets.
\end{itemize}

\subsection{Framework Extensibility}
The architecture is designed for extensibility rather than as a fixed pipeline: the structured-branch encoder could be replaced by any sequence model exposing attention weights; the policy branch could substitute alternative LLMs as they become available; and the lightweight deployment variant could be tuned independently for other low-resource markets beyond CCTS.

\bibliographystyle{IEEEtran}

\end{document}